%% file: main.tex
\documentclass[10pt,twocolumn,letterpaper]{article}
\usepackage[pagenumbers]{cvpr}
\usepackage{enumitem}
\usepackage{amsmath,amssymb}
\usepackage{multirow}

\input{preamble}

\definecolor{cvprblue}{rgb}{0.21,0.49,0.74}
\usepackage[breaklinks,colorlinks,allcolors=cvprblue]{hyperref}

\title{Don’t Miss the Forest for the Trees: Zero-Shot Skeleton Action Recognition with Hyperbolic Hierarchical Semantic Prototypes}
\author{Zixiao Jin\textsuperscript{1,2}\qquad Siyu Zhu\textsuperscript{1,3}\qquad
        Feng Ding\textsuperscript{4}\qquad Haisheng Fu\textsuperscript{5}\\
        Xin Dong\textsuperscript{6}\qquad Shankai Chen\textsuperscript{1,7}\qquad
        Jie Liang\textsuperscript{1}\thanks{Corresponding author.}\\[3pt]
        {\small \textsuperscript{1}Eastern Institute of Technology, Ningbo\qquad
                \textsuperscript{2}China University of Geosciences}\\
        {\small \textsuperscript{3}Xi'an Jiaotong University\qquad
                \textsuperscript{4}Simon Fraser University}\\
        {\small \textsuperscript{5}University of British Columbia\qquad
                \textsuperscript{6}University of Science and Technology of China}\\
        {\small \textsuperscript{7}University of Pennsylvania}}

\hypersetup{
  pdftitle={\thetitle},
  pdfauthor={Zixiao Jin, Siyu Zhu, Feng Ding, Haisheng Fu, Xin Dong, Shankai Chen, Jie Liang}
}

\begin{document}
\maketitle
\input{sec/0_abstract}    
\input{sec/1_intro}
\input{sec/2_related}
\input{sec/3_problem}

\input{sec/4_method}
\input{sec/5_experiment}
\input{sec/6_conclusion}

{
    \small
    \bibliographystyle{ieeenat_fullname}
    \bibliography{main}
}

\end{document}

%% file: preamble.tex
\usepackage{cuted} 

\usepackage{currfile} 

\usepackage{caption} 
\usepackage{colortbl} 
\usepackage{microtype}



%% file: sec/0_abstract.tex
\begin{abstract}
Zero-shot skeleton-based action recognition is evaluated
almost exclusively on flat leaf labels, which leaves open
whether a representation also grounds the broader concepts
that contain an action in skeleton motion. We find that strong
leaf alignment does not carry over to them: when ancestor concepts are scored
directly under independent decoding, eight flat methods with 57.86--73.29\%
mean leaf Top-1 reach only 6.15--31.21\% mean full-path accuracy across four
NTU RGB+D splits. We propose the \textbf{H}yperbolic
\textbf{H}ierarchical \textbf{S}emantic \textbf{P}rototype ($H^2SP$) framework, which represents every
node of an action taxonomy in the Lorentz model of hyperbolic space. First,
instead of learning a free vector per node, $H^2SP$ synthesizes
hierarchy-aware prototypes by recursive parent--child geodesic interpolation
under a single tree-wide gate. Second, tangent-space semantic and
distributional alignment on seen classes anchor the leaf prototypes in
skeleton motion. Third, a soft entailment-cone regularizer encourages
directional parent--child inclusion, propagating this motion grounding to
ancestor concepts. We evaluate hierarchical concept prediction under
independent and cascade decoding, and transfer to an alternative taxonomy
without retraining. Across the four splits, $H^2SP$ attains the best mean
independent and cascade full-path accuracy among the
evaluated methods, at a comparable leaf Top-1 of 72.40\%. Transferred to the alternative taxonomy, it outperforms both flat baselines and the same-tree Euclidean control across every split and decoding mode, indicating that its concept-level correspondence extends beyond the training taxonomy.
\looseness=-1\par
\end{abstract}

%% file: sec/1_intro.tex
\section{Introduction}
\label{sec:intro}

\begin{figure}[t]
    \centering
    \includegraphics[width=\linewidth]{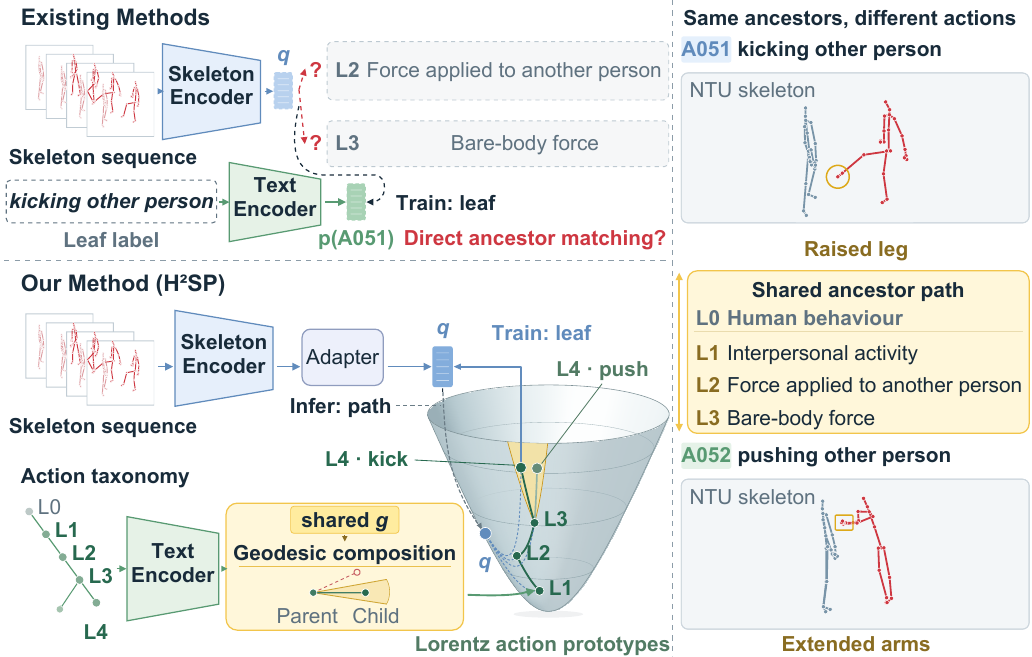}
 \caption{\textbf{Leaf labels vs.\ the concepts that contain them.}
 \emph{Top-left:} flat methods align the skeleton query $q$ only with leaf
 labels (\emph{kicking other person}), leaving direct matching with ancestor descriptions ($L_2$, $L_3$) untested by leaf-level evaluation. \emph{Right:} \emph{kicking} and \emph{pushing other person} share
 the ancestor path $L_0$--$L_3$ and differ only in leaf-level joint cues
 (raised leg vs.\ extended arms). \emph{Bottom-left:} $H^2SP$ composes Lorentz
 prototypes for every node by gated geodesic interpolation and entailment
 cones, without free per-node vectors, so $q$ can be matched at any level.}
 \label{fig:teaser}
\end{figure}

Zero-shot skeleton-based action recognition transfers motion--language
correspondences learned on seen classes to classes absent from training.
Recent methods strengthen this transfer with contextual side information and
evolving prototypes~\cite{neuron}, frequency-aware semantic
alignment~\cite{fsvae}, diffusion-based modality
alignment~\cite{tdsm}, and neighbor-aware
semantics~\cite{flora}. Their evaluation, however, asks a single
question: which unseen \emph{leaf} label is closest to the motion? A skeleton of one person kicking another is labeled \textit{kicking other person}, yet it is also an instance of \textit{force applied to another person} alongside \textit{pushing other person} (Fig.~\ref{fig:teaser}). Querying such a shared concept asks whether the skeleton representation matches that concept’s description directly, rather than inferring the ancestor indirectly after selecting a leaf label. Closed leaf lookup bypasses this direct matching, and leaf Top-1 accuracy alone cannot test whether broader concepts are grounded across taxonomy levels.
 
The gap is measurable. One might expect a text encoder aligned with skeletons
at the leaves to already place broader descriptions near the right motions,
since language models capture some hypernymy. The evidence says otherwise.
When broader concepts are scored directly under independent decoding, eight
recent flat methods~\cite{smie,sadvae,fsvae,tdsm,dvta,flora,neuron,pgfa}
that reach 57.86--73.29\% mean leaf Top-1 on the NTU RGB+D splits recover the
complete concept path for only 6.15--31.21\% of test samples
(Sec.~\ref{sec:main}). The pattern holds even for the most accurate leaf
classifier, FLORA~\cite{flora} (73.29\% leaf Top-1, 6.15\% full path). Because these methods optimize leaf discrimination, we directly evaluate skeleton-to-concept matching at every taxonomy level rather than relying on leaf-to-ancestor lookup.
 
We propose \textbf{\boldmath$H^2SP$}, a Hyperbolic Hierarchical Semantic Prototype
framework that gives every taxonomy node, not only the leaves, a prototype in
the Lorentz model of hyperbolic space, whose exponential volume growth suits
trees~\cite{lorentz}. Unlike prior hyperbolic action models that
learn concept embeddings from video for hierarchical
retrieval~\cite{hyperbolic_action} or unify datasets under a shared
taxonomy~\cite{pangea}, $H^2SP$ grounds a hierarchy in sparse skeleton
motion under a zero-shot seen/unseen split.
 
The central difficulty is that motion supervision exists only for seen
leaves, while every ancestor must still be recognizable from motion. $H^2SP$
connects the two in three steps. \emph{First}, prototypes are composed, not
stored: each node moves from its parent's composed prototype toward its own
text seed along a geodesic, with one learned gate shared by the whole tree.
No free per-node vectors are optimized, so seen and unseen leaves inherit the
same ancestor context, and a concept shared by several actions has a single
prototype. \emph{Second}, in seen classes, the semantic and distributional alignments in tangent space tie each leaf prototype to the skeletons of its
class, anchoring the hierarchy in motion. \emph{Third}, a soft entailment-cone
regularizer~\cite{cones} encourages each child to lie inside
its parent's cone, so the grounding obtained at the leaves propagates upward
to their ancestors.
 
We evaluate \emph{hierarchical concept prediction} under two decoding modes.
\emph{Independent} decoding matches the query against all candidates at each
level. \emph{Cascade} decoding descends the tree, restricting each level to the
children of the previous prediction. Because $H^2SP$ uses the taxonomy during
training and flat baselines do not, we separate three comparisons:
(i)~flat baselines scored on the same candidate descriptions at every level,
each through its own text--skeleton pipeline, with the same tree links for
cascade routing, which tests whether leaf alignment alone preserves broader
concepts; (ii)~transfer, without retraining, to an alternative taxonomy whose
internal concepts and links are unseen by every method; and
(iii)~Euclidean counterparts trained with the same tree and composition, with
and without cone loss.
 
\noindent Our contributions are threefold:

\noindent \textbf{\boldmath$H^2SP$.} We introduce a Lorentz prototype generator that composes a prototype for every taxonomy node by geodesic interpolation with a single tree-wide gate, without free per-node vectors. Semantic and distributional alignment ground seen-leaf prototypes in skeleton motion, while soft entailment-cone regularization encourages parent--child inclusion.

\noindent \textbf{Hierarchical zero-shot evaluation.} We evaluate concept matching independently at each taxonomy level and under tree-constrained cascade decoding. Transfer to a supplied alternative taxonomy tests internal concepts and parent links unseen during training, with model parameters and source leaf scores fixed.

\noindent \textbf{Experimental evidence.} In four NTU RGB+D 60/120 splits, H$^{2}$SP achieves the highest mean independent and cascade full-path accuracies among the evaluated methods, at 43.46\% and 50.64\%. Its mean leaf Top-1 remains competitive at 72.40\%. Its full-path advantage over all transfer baselines persists on every split under the alternative taxonomy. Ablations on NTU60~55/5 and NTU120~96/24 support the contributions of seen-leaf alignment, shared gating, and the coupling of Lorentz geometry with cone regularization.

%% file: sec/2_related.tex
\section{Related Work}
\label{sec:related}

\noindent\textbf{Skeleton-based action recognition.}
Graph convolutional networks dominate skeleton modeling, from
ST-GCN~\cite{stgcn} to adaptive and channel-wise
topologies~\cite{2sagcn,ctrgcn}, multi-scale aggregation~\cite{msg3d}, shift
graph operations~\cite{shiftgcn}, information bottlenecks~\cite{infogcn}, and
hierarchically decomposed joint graphs~\cite{hdgcn}; PoseC3D~\cite{posec3d}
uses 3D heatmaps and GAP~\cite{lst} supervises the encoder with
LLM-generated part descriptions. They recognize closed label sets, typically
on NTU RGB+D~\cite{ntu60,ntu120}, and any hierarchy they model is over joints,
not action concepts. As in prior zero-shot work~\cite{synse,smie,tdsm,flora},
we use fixed features from a skeleton encoder trained on seen classes.

\noindent\textbf{Zero-shot skeleton action recognition.}
Early methods learn a joint pose--language space~\cite{jpose} or align
modalities via generative models~\cite{synse,msf,sadvae} or mutual
information~\cite{smie}. Recent work enriches text with LLM part-level
descriptions~\cite{purls,star,dynapurls}, multi-granularity
semantics~\cite{infocomp}, contextual evolving prototypes~\cite{neuron}, or
frequency-aware semantics~\cite{fsvae}, and improves matching through dual
alignment~\cite{dvta}, prototype guidance~\cite{pgfa},
diffusion~\cite{tdsm}, flow matching~\cite{protofm}, or open-form
flows~\cite{flora}; open-vocabulary recognition across skeleton formats has
also been studied~\cite{universal}. Multi-granularity text in these methods
describes the same leaf more richly, and evaluation still selects one unseen
leaf. We make each leaf's ancestors separate prediction targets.
\looseness=-1\par

\noindent\textbf{Class hierarchies in recognition.}
Zero-shot recognition has long used class structure through label-name
embeddings~\cite{devise}, attribute or hierarchy-based label
embeddings~\cite{labelembed}, and knowledge-base relatedness~\cite{rohrbach}.
Hierarchies also reduce mistake severity~\cite{hierloss}, organize
fine-grained video benchmarks~\cite{finegym}, and guide coarse features in
supervised skeleton recognition~\cite{hiod}. There, the hierarchy improves
leaf prediction or is learned with labels at every level. In our zero-shot
setting, internal concepts span unseen children, and at test time may come
from a taxonomy never seen in training. We therefore apply motion
supervision only at seen leaves and recognize internal concepts from their
descriptions, rather than training a fixed classifier per ancestor on seen
samples.

\noindent\textbf{Hyperbolic representation learning.}
Hyperbolic space embeds trees with low distortion. Poincar\'e~\cite{poincare}
and Lorentz~\cite{lorentz} embeddings, Lorentzian distances~\cite{lorentzdist},
order embeddings~\cite{vendrov}, and entailment cones~\cite{cones} encode
hierarchies and partial orders, and hyperbolic layers~\cite{hnn,hgcn,fullyhyp}
make them trainable. Vision applications include image
embeddings~\cite{khrulkov}, metric learning~\cite{hyptransfer},
segmentation~\cite{atigh}, CLIP-style~\cite{clip} image--text models with
entailment objectives~\cite{meru,hycoclip}, and compositional zero-shot
learning~\cite{li2025learning}. For actions, hyperbolic models predict the
future at an adaptive level of an action hierarchy~\cite{hyperfuture},
retrieve related actions with prototypes learned for every
node~\cite{hyperbolic_action}, and unify heterogeneous datasets in a
text-derived Lorentz label space~\cite{pangea}. For skeletons,
HYSP~\cite{hysp} paces self-supervised learning with hyperbolic uncertainty,
and SkelHCC~\cite{skelhcc} models a joint--part--body hierarchy for one-shot
recognition. $H^2SP$ differs from both lines: its hierarchy is a semantic
action taxonomy, and its node prototypes are composed by shared geodesic
interpolation rather than stored per node, with skeleton grounding at the
leaves carried upward by entailment cones.
\looseness=-1\par

%% file: sec/3_problem.tex
\section{Problem Formulation}
\label{sec:problem}

\noindent\textbf{Setting and targets.}
Let $\mathcal{S}$ and $\mathcal{U}$ be disjoint sets of seen and unseen action
classes, with training data $\mathcal{D}_S=\{(X_i,y_i):y_i\in\mathcal{S}\}$.
We predict unseen leaves and their containing concepts from the same skeleton. A rooted
taxonomy $\mathcal{T}=(\mathcal{V},\mathcal{E})$ has leaves
$\mathcal{S}\cup\mathcal{U}$; every node $v$ has a text description $d_v$ and,
except the root, a unique parent $\mathrm{pa}(v)$. With $L$ edges from root
to leaf, the target path of class $a$ is
\begin{equation}
\pi(a)=(v_{a,0},v_{a,1},\ldots,v_{a,L}),
\label{eq:path}
\end{equation}
from root $v_{a,0}$ to leaf $v_{a,L}$, with taxonomy-defined depth $L$.
The candidates at level $k$ are the distinct ancestors of the
unseen leaves,
\begin{equation}
\mathcal{V}_k(\mathcal{U})=\{v_{a,k}:a\in\mathcal{U}\},\qquad k=1,\ldots,L,
\label{eq:cand}
\end{equation}
so each shared internal concept appears once. Specifically, $\mathcal V_k(\mathcal U)$ deduplicates level-$k$ ancestors induced by $\mathcal U$, including internal concepts shared with $\mathcal S$.

\noindent\textbf{Direct concept prediction.}
\emph{Independent} decoding predicts one node from $\mathcal{V}_k(\mathcal{U})$
at each level without conditioning on other levels. \emph{Cascade} decoding
starts at the root and scores only children of the previous prediction
$\hat v^{\mathrm{cas}}_{k-1}$, committing subsequent predictions to that
subtree without backtracking. $H^2SP$ scoring rules appear in
Sec.~\ref{sec:infer}. On test set $\mathcal{Q}=\{(X_i,a_i)\}_{i=1}^{N}$,
exact full-path accuracy under either mode, excluding the shared root, is
\begin{equation}
A_{\mathrm{full}}=\frac1N\sum_{i=1}^{N}\prod_{k=1}^{L}
\mathbf{1}\big[\hat v_{i,k}=v_{a_i,k}\big],
\label{eq:full}
\end{equation}
reported with independent leaf Top-1 accuracy ($\hat v^{\mathrm{ind}}_L$). 

On a fixed taxonomy with a unique root-to-leaf path, predicting a leaf and looking up its ancestors yields an exact full-path accuracy equal to leaf Top-1 accuracy. While valid as an indirect readout when all target leaves are enumerated, leaf lookup does not separately evaluate compatibility between the skeleton query and internal-node descriptions (Fig.~\ref{fig:teaser}). We therefore score candidate concepts directly at each level. Independent $A_{\mathrm{full}}$ cannot exceed leaf Top-1 accuracy, since a correct full path requires a correct leaf prediction.
\looseness=-1\par

\noindent\textbf{Cross-taxonomy transfer.}
A model trained with taxonomy $\mathcal{T}_A$ is evaluated on an alternative
taxonomy $\mathcal{T}_B$ over the same leaves, whose internal concepts and
parent links are supplied only at inference; depths $L_A$ and $L_B$ may
differ. Parameters, training statistics, and source leaf scores stay frozen;
only target internal scores and path decoding are recomputed
(Sec.~\ref{sec:infer}), isolating transfer at internal levels.

%% file: sec/4_method.tex
\section{Method}
\label{sec:method}

\begin{figure*}[t]
  \centering
  \includegraphics[width=\linewidth,trim=0 4bp 0 4.5bp,clip]{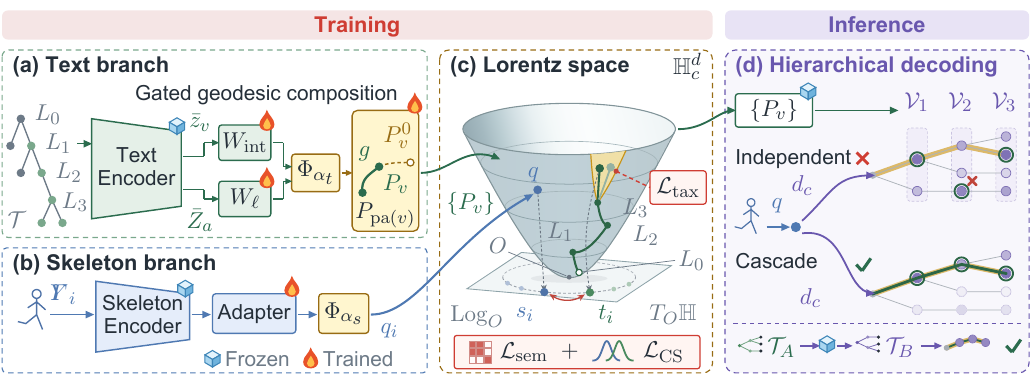}
  \caption{\textbf{\boldmath Schematic overview of $H^2SP$.} (a)~Frozen text-encoder features of the
  taxonomy nodes are projected and lifted to Lorentz seeds $P^0_v$;
  parent--seed geodesic composition uses shared gate $g$ to obtain $P_v$.
  (b)~A frozen skeleton encoder and a residual adapter give the query $q_i$.
  (c)~Each loss acts where it is drawn ($\mathcal L_{\mathrm{tax}}$ on the cone,
  $\mathcal L_{\mathrm{sem}}$ and $\mathcal L_{\mathrm{CS}}$ on $s_i,t_i$ at $O$).
  (d)~An unseen query is scored against the frozen prototypes of
  $\mathcal V_k(\mathcal U)$ by $d_c$, per level or cascaded; a sample is
  correct only if the whole path is.}
  \label{fig:framework}
\end{figure*}

\noindent\textbf{Overview.}
$H^2SP$ (Fig.~\ref{fig:framework}) builds a prototype for every taxonomy node
in Lorentz space (Sec.~\ref{sec:compose}), aligns skeleton queries with the
composed leaf prototypes on seen classes and regularizes parent--child
geometry with entailment cones (Sec.~\ref{sec:align}), and matches the same
query against candidate concepts at every level at test time
(Sec.~\ref{sec:infer}). Every operator is shared across the supplied tree, so
depth is a property of the taxonomy rather than of the architecture.

\subsection{Lorentz Preliminaries}
\label{sec:lorentz}
For curvature $-c$ ($c>0$), the $d$-dimensional Lorentz model and its inner
product are~\cite{lorentz}
\begin{equation}
\begin{aligned}
\mathbb{H}^d_c&=\{p\in\mathbb{R}^{d+1}:\langle p,p\rangle_{\mathcal L}=-1/c,\ p_t>0\},\\
\langle p,q\rangle_{\mathcal L}&=p_s^\top q_s-p_tq_t,
\end{aligned}
\label{eq:lorentz}
\end{equation}
with spatial part $p_s\in\mathbb{R}^d$ and time-like coordinate $p_t$, and the
geodesic distance is
\begin{equation}
d_c(p,q)=\tfrac{1}{\sqrt c}\,\mathrm{arcosh}\big(-c\langle p,q\rangle_{\mathcal L}\big).
\label{eq:dist}
\end{equation}
The origin is $O=(\mathbf{0},c^{-1/2})$, and $\mathrm{Exp}^c_p$, $\mathrm{Log}^c_p$
denote the exponential and logarithmic maps at $p$, and
$\|v\|_{\mathcal L}=\sqrt{\langle v,v\rangle_{\mathcal L}}$ is the norm of a
tangent vector~\cite{lorentz,meru}. For
stability we use a capped map
$\mathrm{Exp}^c_{p,R}(v)=\mathrm{Exp}^c_p(\min\{1,R/(\sqrt c\,\|v\|_{\mathcal L})\}\,v)$
with a fixed cap $R$ (the multiplier is 1 for $v=0$), and lift a Euclidean vector $h\in\mathbb{R}^d$ by
$\Phi_\alpha(h)=\mathrm{Exp}^c_{O,R}((\alpha h,0))$ with a modality-specific
scale $\alpha$. Closed forms and numerical safeguards are in the supplementary
material.

\subsection{Dynamic Hierarchical Prototype Construction}
\label{sec:compose}
\noindent\textbf{Text seeds.}
Leaves reuse FLORA's frozen dual-description CLIP ViT-L/14@336px
features~\cite{flora,clip}, concatenated into
$Z_a\in\mathbb{R}^{T\times1536}$ with the pooled features at position $T$.
Internal nodes, including the root, have no such cache and use one pooled
feature $z_v\in\mathbb{R}^{768}$ from the same CLIP model. Internal features are standardized with statistics of the nodes on
seen-class paths and leaf features with those of seen classes; both stay
fixed at inference.

With $d=256$, one projection is shared by all internal nodes and
another by all leaves:
\begin{equation}
h_v=W_{\mathrm{int}}\bar z_v+b_{\mathrm{int}},\qquad
A_{a,t}=W_\ell\bar Z_{a,t}+b_\ell .
\label{eq:seed}
\end{equation}
The pooled leaf token queries all tokens through scaled dot-product
attention~\cite{attention},
\begin{equation}
\omega_{a,t}=\frac{\exp(A_{a,t}^\top A_{a,T}/\sqrt d)}
{\sum_{j=1}^{T}\exp(A_{a,j}^\top A_{a,T}/\sqrt d)},\quad
h^\ell_a=\sum_{t=1}^{T}\omega_{a,t}A_{a,t},
\label{eq:attn}
\end{equation}
with queries, keys, and values all from $W_\ell$, so the readout adds no
class-specific text parameters.

\noindent\textbf{Parent-conditioned geodesic composition.}
Each node $v$ (with $h_v=h^\ell_a$ for a leaf) starts from
$P^0_v=\Phi_{\alpha_t}(h_v)$, and the root keeps $P_{\mathrm{root}}=P^0_{\mathrm{root}}$.
Every other node moves from its composed parent toward its own seed along
their geodesic,
\begin{equation}
P_v=\mathrm{Exp}^c_{P_{\mathrm{pa}(v)},R}\big(g\,\mathrm{Log}^c_{P_{\mathrm{pa}(v)}}(P^0_v)\big),\quad g=\mathrm{sigmoid}(\gamma),
\label{eq:compose}
\end{equation}
where $\gamma$ is one learned scalar shared by all edges. The parent fixes the
context and the node's text fixes the displacement; applying
Eq.~\ref{eq:compose} in topological order carries information from broad
concepts down to their descendants, and the leaf prototype is
$p_a=P_{v_{a,L}}$. Prototypes are recomputed from frozen text features and
shared parameters at every forward pass instead of being stored, so seen and
unseen leaves use the same mapping, a shared ancestor has a single prototype
for all its descendants, and a new taxonomy can be encoded without retraining.

\subsection{Cross-Modal Alignment and Hierarchy Regularization}
\label{sec:align}
Composition links the concepts to one another but not to motion. We
therefore train a residual skeleton adapter and align its queries with the
composed leaf prototypes of seen classes.
\looseness=-1\par

\noindent\textbf{Skeleton queries.}
A frozen Shift-GCN~\cite{shiftgcn} gives $x\in\mathbb{R}^d$, which a residual
adapter corrects before lifting,
\begin{equation}
\tilde x=x+r\,W_u\,\mathrm{GELU}\big(W_d\,\mathrm{LN}(x)+b_d\big),\qquad q=\Phi_{\alpha_s}(\tilde x),
\label{eq:adapter}
\end{equation}
with $W_d,W_u\in\mathbb{R}^{d\times d}$, $b_d\in\mathbb{R}^d$, and a fixed
residual scale $r$. $\tilde x$ is not unit-normalized, so $q$ keeps a radius
(capped at $R/\sqrt c$) that enters $d_c$ at inference.

\noindent\textbf{Semantic alignment.}
For a mini-batch $\{(x_i,y_i)\}_{i=1}^{B}$, queries and their leaf prototypes
are mapped to the tangent space at $O$ and their spatial parts
$\ell_2$-normalized by $\mathcal N$,
$s_i=\mathcal N([\mathrm{Log}^c_O(q_i)]_s)$ and
$t_i=\mathcal N([\mathrm{Log}^c_O(p_{y_i})]_s)$, so both alignment terms
constrain directions only; prototype radii are shaped by composition and the cone term. With logits $\ell_{ij}=s_i^\top t_j/\tau$ and temperature $\tau>0$, we adapt symmetric InfoNCE~\cite{cpc,clip,csaligner} to class-level supervision by treating all same-label entries as positives:
\begin{equation}
\begin{aligned}
\mathcal I(\ell)&=-\frac1B\sum_{i=1}^{B}\log
\frac{\sum_{j:y_j=y_i}\exp(\ell_{ij})}{\sum_{j=1}^{B}\exp(\ell_{ij})},\\
\mathcal L_{\mathrm{sem}}&=\tfrac12\big(\mathcal I(\ell)+\mathcal I(\ell^\top)\big).
\end{aligned}
\label{eq:sem}
\end{equation}
Same-class samples are thus never treated as negatives.

\noindent\textbf{Distributional alignment.}
Contrastive objectives alone can leave the two modalities separated by a
global offset in the shared space, and reducing this gap has been tied to
robustness~\cite{gap}. We therefore also match the empirical distributions
with the Gaussian-kernel Cauchy--Schwarz divergence of
CS-Aligner~\cite{csaligner} on the same tangent features. For the batch
banks $\mathcal B_s=(s_i)_{i=1}^{B}$ and $\mathcal B_t=(t_i)_{i=1}^{B}$, kernel
$K_\sigma(u,v)=\exp\!\big(-(2-2u^\top v)/(2\sigma^2)\big)$ with bandwidth
$\sigma>0$, and mean kernel
$\bar K(U,V)=\frac{1}{B^2}\sum_{i,j=1}^{B}K_\sigma(u_i,v_j)$ for
$U,V\in\{\mathcal B_s,\mathcal B_t\}$ over all ordered pairs, including
self-pairs,
\begin{equation}
\begin{aligned}
\mathcal L_{\mathrm{CS}}={}&\log\bar K(\mathcal B_s,\mathcal B_s)+\log\bar K(\mathcal B_t,\mathcal B_t)\\
&-2\log\bar K(\mathcal B_s,\mathcal B_t).
\end{aligned}
\label{eq:cs}
\end{equation}

\noindent\textbf{Entailment-cone regularization.}
Composition passes ancestor information down but does not make the parent
contain the child. Following the taxonomic entailment regularization of
H$^2$em~\cite{li2025learning}, we penalize children outside their parents'
entailment cones~\cite{cones}, in the Lorentz form of MERU~\cite{meru}. For a
parent $p\neq O$ and child $q\neq p$, with $m=-c\langle p,q\rangle_{\mathcal L}$,
the exterior angle and the cone half-aperture are
\begin{equation}
\begin{aligned}
\Theta(p,q)&=\arccos\Big(\frac{q_t-m\,p_t}{\|p_s\|_2\sqrt{m^2-1}}\Big),\\
\Psi(p)&=\arcsin\Big(\min\Big\{1,\frac{2K}{\sqrt c\,\|p_s\|_2}\Big\}\Big),
\end{aligned}
\label{eq:cone}
\end{equation}
with aperture constant $K>0$. Over the set $\mathcal E_B$ of distinct
parent--child edges on the paths $\pi(y_i)$ of the mini-batch classes,
\begin{equation}
\mathcal L_{\mathrm{tax}}=\frac1{|\mathcal E_B|}\sum_{(u,v)\in\mathcal E_B}\big[\Theta(P_u,P_v)-\Psi(P_u)\big]_+ .
\label{eq:tax}
\end{equation}
Broader concepts act as cone apices and their descendants are encouraged to
stay inside. Since every leaf prototype is composed through its ancestors
(Eq.~\ref{eq:compose}), leaf alignment updates the shared projections and
gate that also produce the ancestors, and the cone term constrains these
ancestors directly, without any motion label at internal nodes. It is a soft
penalty and does not by itself guarantee a strict partial order.

\noindent\textbf{Joint objective.}
The adapter, the two text projections, and the gate are trained with
\begin{equation}
\mathcal L=\lambda_{\mathrm{align}}\big(\mathcal L_{\mathrm{sem}}+\lambda_{\mathrm{CS}}\mathcal L_{\mathrm{CS}}\big)+\lambda_{\mathrm{tax}}\mathcal L_{\mathrm{tax}},
\label{eq:joint}
\end{equation}
with $\lambda_{\mathrm{align}},\lambda_{\mathrm{CS}},\lambda_{\mathrm{tax}}\ge0$;
encoders, curvature, modality scales, temperature, $R$, and $K$ are fixed
within a run. As in flat baselines, motion supervision acts only between
skeleton queries and seen leaf prototypes. The taxonomy enters as
class-level structure, which flat baselines do not receive; Sec.~\ref{sec:abl}
therefore also reports Euclidean controls trained with the same tree.

\subsection{Hierarchical Inference}
\label{sec:infer}
Independent decoding scores every candidate in $\mathcal V_k(\mathcal U)$
(Eq.~\ref{eq:cand}) at each level by geodesic distance, and cascade decoding
fixes $\hat v^{\mathrm{cas}}_0$ to the root and descends without backtracking:
\begin{align}
\hat v^{\mathrm{ind}}_k&=\arg\min_{v\in\mathcal V_k(\mathcal U)}d_c(q,P_v),\label{eq:ind}\\
\hat v^{\mathrm{cas}}_k&=\arg\min_{\substack{v\in\mathcal V_k(\mathcal U)\\ \mathrm{pa}(v)=\hat v^{\mathrm{cas}}_{k-1}}}d_c(q,P_v),\label{eq:cas}
\end{align}
for $k=1,\ldots,L$. Both modes are scored with Eq.~\ref{eq:full}, and leaf
Top-1 is the level-$L$ independent prediction.

\noindent\textbf{Transfer to an alternative taxonomy.}
We freeze the checkpoint and all seen-class statistics learned on
$\mathcal T_A$. The internal descriptions and parent links of $\mathcal T_B$
are encoded by the same shared construction (Eqs.~\ref{eq:seed}
and~\ref{eq:compose}). The source leaf score matrix is retained, and only
target internal scores and path decoding are recomputed, which isolates
internal-concept transfer under a taxonomy shift without retraining or
re-encoding target leaves.

%% file: sec/5_experiment.tex
\section{Experiments}
\label{sec:exp}
 
\input{sec/tables/main_results_combined}

 \begin{figure*}[!t]
    \centering
    \includegraphics[width=\linewidth,trim=0 0.5bp 0 1.5bp,clip]{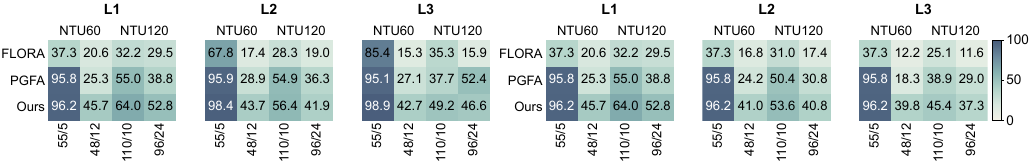}

    \caption{\textbf{Internal concept accuracy ($L_1$--$L_3$) under independent (left) and cascade (right) decoding.} FLORA degrades sharply toward coarser levels ($85.4\%\!\rightarrow\!37.3\%$ from $L_3$ to $L_1$ on NTU60 55/5), whereas $H^2SP$ leads on 11/12 (independent) and 12/12 (cascade) split--level pairs.}
    \label{fig:level_accuracy}

\end{figure*}

\begin{figure*}[!t]
    \centering
    \begin{minipage}[c]{0.62\linewidth}
        \centering
        \includegraphics[width=\linewidth]{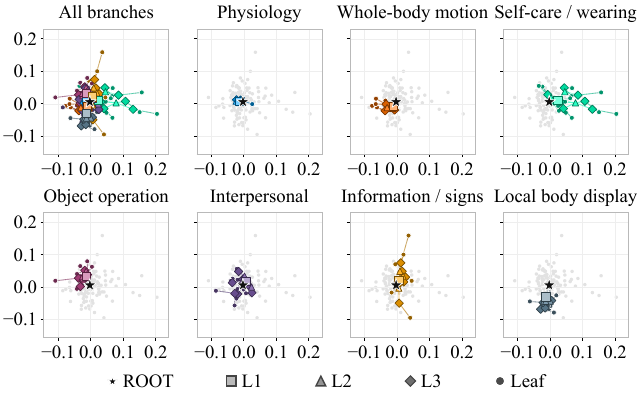}
        \caption{Projected geometry of the 127 composed prototypes on NTU60. }
        \label{fig:prototype_branches}
    \end{minipage}\hfill
    \begin{minipage}[c]{0.33\linewidth}
        \centering
        \includegraphics[width=\linewidth,trim=0 0.5bp 0 1.5bp,clip]{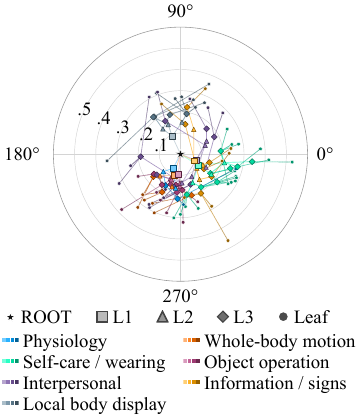}
        \caption{Manifold root distance $r_v=d_c(P_{\mathrm{root}},P_v)$ by taxonomy level.}
        \label{fig:prototype_radius}
    \end{minipage}

\end{figure*}

\begin{figure*}[!t]
    \centering
    \includegraphics[width=\linewidth,trim=0 0.75bp 0 1.5bp,clip]{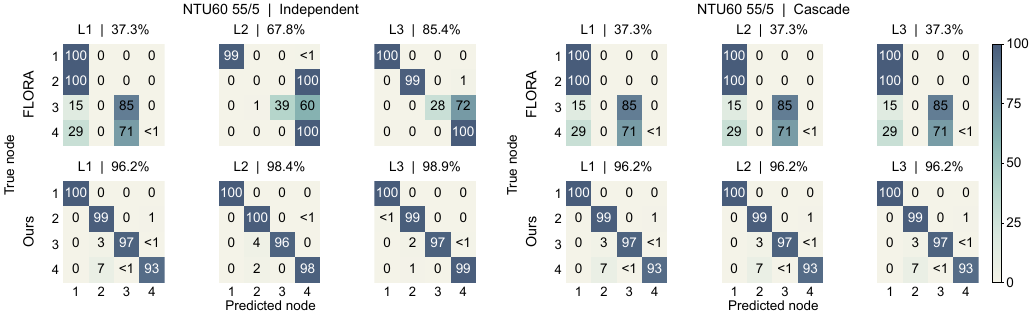}
    \caption{\textbf{Row-normalized internal-node confusion matrices ($L_1$--$L_3$) on NTU60 55/5 under independent (left) and cascade (right) decoding.} FLORA collapses onto nodes 1 and 3 at $L_1$ (37.3\%) and locks cascade predictions into the same error pattern across $L_2$--$L_3$, whereas $H^2SP$ maintains 93\%--100\% diagonal recall.}
    \label{fig:confusion_matrices}
\end{figure*}
\subsection{Setup}
\label{sec:setup}
\noindent\textbf{Data and taxonomies.}
We use NTU RGB+D 60 and 120~\cite{ntu60,ntu120} under the standard splits 55/5, 48/12, 110/10, and 96/24~\cite{synse,flora,tdsm}. Over identical flat official leaves, we define two frozen, rule-based class-to-path taxonomies (not per-video attribute annotations): (i) $\mathcal T_A$ ($L{=}4$: $\text{ROOT}\!\to\!L_1\!\to\!L_2\!\to\!L_3\!\to\!\text{Leaf}$) maps \emph{semantic domain}$\to$\emph{event family}$\to$\emph{fine concept}$\to$\emph{action}~\cite{bach1968universals,baker1998berkeley,kipper2008verbnet}; (ii) $\mathcal T_B$ ($L{=}3$: $\text{ROOT}\!\to\!L_1\!\to\!L_2\!\to\!\text{Leaf}$) splits $L_1$ by \emph{qualitative relation change and repetition} and $L_2$ by \emph{relation-change combinations}~\cite{aksoy2011learning,zacks2001event} via a 7-bit physical signature (e.g., contact, support, wearing, object pose). For instance, \texttt{A012} (\emph{writing}) maps to \text{ROOT} $\to$ \textrm{Information \& symbolic activity} $\to$ \textrm{Information reading \& input} $\to$ \textrm{Text reading/writing \& input} $\to$ \text{writing} ($\mathcal T_A$) and \text{ROOT} $\to$ \textrm{Repetitive motion with qualitative relation change} $\to$ \textrm{Contact, external-object pose} $\to$ \text{writing} ($\mathcal T_B$). Across splits, $|\mathcal V_k(\mathcal U)|$ ($\text{ROOT}\!\to\!\text{Leaf}$) is $1/4/4/4/5$, $1/6/10/11/12$, $1/6/9/10/10$, $1/5/13/19/24$ on $\mathcal T_A$, and $1/3/5/5$, $1/3/8/12$, $1/3/8/10$, $1/4/10/24$ on $\mathcal T_B$; full rules, bit definitions, more path examples, and trees are in the supplementary material.
 
\noindent\textbf{Implementation and protocol.}
Skeleton features come from a frozen Shift-GCN~\cite{shiftgcn} and text
features from CLIP ViT-L/14@336px (Sec.~\ref{sec:compose}), with $d=256$.
For $H^2SP$, we hold out approximately 10\% of the seen classes as
class-disjoint pseudo-unseen validation classes and train on the remaining
seen classes. Hyperparameters are selected by leaf Top-1 accuracy on this
validation set; the official unseen classes are reserved for final
evaluation. Additional implementation details are provided in the
supplementary material.
 \looseness=-1\par
 
\noindent\textbf{Comparison tracks.}
All methods see the same sample-level supervision $\mathcal D_S$, without
skeleton-to-ancestor labels. In the \emph{native track} (Sec.~\ref{sec:main}),
published flat methods keep their own training and class-level priors; each
scores the shared candidates $\mathcal V_k(\mathcal U)$ through its own
text--skeleton pipeline and receives the same parent links for cascade
routing. Because $H^2SP$ alone also trains with $\mathcal T_A$, this track
tests whether strong leaf alignment by itself preserves broader concepts, not
which method wins under identical supervision. Transfer to $\mathcal T_B$ evaluates unseen internal concepts (Sec.~\ref{sec:xtax}), while Euclidean variants provide geometry controls under the same taxonomy supervision (Sec.~\ref{sec:abl}).
 
\subsection{Main Results}
\label{sec:main}

$H^2SP$ leads both full-path metrics on all four splits
(Table~\ref{tab:main_results}(a,b)), averaging 43.46\% independent and 50.64\%
cascade accuracy. Its margins over the best standard inductive baselines
are 22.64 points (SA-DVAE, 20.82\%) and 16.12 points (DVTA, 34.52\%),
and over PGFA$^\dagger$ (31.21\%/40.43\%) are 12.25/10.21 points.
Mean leaf Top-1 is 72.40\%, compared with FLORA's 73.29\%; $H^2SP$ ranks
second on NTU120 110/10 (76.92\%) and 96/24 (67.41\%, behind TDSM's
67.84\%). Despite its leaf accuracy, FLORA reaches only 6.15\% independent
full-path accuracy, including 0.00\% on NTU60 48/12.
This contrast probes direct ancestor matching beyond leaf classification
(Sec.~\ref{sec:problem}). Protocols remain explicit: PGFA$^\dagger$ uses
test-time adaptation, while Neuron$^\ddagger$ reports a candidate-set hit
rate on a filtered subset.
 



\noindent\textbf{Level-wise breakdown.}
Fig.~\ref{fig:level_accuracy} shows where full paths fail. Under independent decoding,
FLORA is close to its leaf accuracy at $L_3$ on NTU60 55/5 (85.4\% vs.\
86.28\%) but drops toward coarser levels (67.8\% at $L_2$, 37.3\% at $L_1$),
and stays below 35.5\% at every internal level on the other splits.
$H^2SP$ reaches 96.2/98.4/98.9\% at $L_1$/$L_2$/$L_3$ on NTU60 55/5 and leads
in 11 of 12 split--level pairs; the exception is $L_3$ on NTU120 96/24
(PGFA$^\dagger$ 52.4\% vs.\ 46.6\%), where $H^2SP$ still leads at $L_1$ and
$L_2$. Under cascade decoding it leads in all 12 pairs.
 
\subsection{Cross-Taxonomy Generalization}
\label{sec:xtax}
We replace the internal descriptions and parent links of $\mathcal{T}_A$ by those of $\mathcal{T}_B$ at inference, keeping weights, source statistics, and source leaf scores fixed (Sec.~\ref{sec:problem}).

\input{sec/tables/cross_taxonomy}

We compare $H^2SP$ against the top flat baselines from
Table~\ref{tab:main_results} (FLORA and PGFA) as well as a
\textbf{Euclidean + Cone} control (Sec.~\ref{sec:abl})
trained on the same data and taxonomy $\mathcal{T}_A$
while retaining gated composition and cone regularization,
with hyperparameters tuned separately by leaf Top-1 accuracy
on the same pseudo-unseen validation classes.
As shown in Table~\ref{tab:cross_tree}, $H^2SP$ ranks first
on every split under both decoders, averaging
$11.67\%/17.40\%$ (independent/cascade) vs.\
$9.92\%/13.08\%$ for Euclidean + Cone and
$2.34\%/11.14\%$ for FLORA.
Two comparisons emerge.
First, even in Euclidean space, the Euclidean + Cone control
reaches $9.92\%$ mean independent full-path accuracy,
substantially outperforming FLORA ($2.34\%$) and PGFA ($1.20\%$)
on the target taxonomy.
Second, with cone regularization included in both hierarchical
models and hyperparameters selected under the same validation
criterion, $H^2SP$ improves over Euclidean + Cone by
$1.75/4.33$ points in mean independent/cascade full-path
accuracy (and by $7.27$ cascade points on NTU120~96/24),
supporting the advantage of the Lorentz-based formulation
on the supplied target taxonomy.
\looseness=-1\par

\subsection{Decoding Diagnostics and Qualitative Case}
\label{sec:tradeoff}


Fig.~\ref{fig:confusion_matrices} (left) shows FLORA collapsing onto a few internal nodes: at $L_1$, true nodes 2 (\textit{Self-care / wearing}) and 4 (\textit{Information / signs}) are absorbed into nodes 1 and 3 ($<1\%$ recall). Because 55/5 candidates form four single-child chains across $L_1$--$L_3$, cascade decoding repeats this $L_1$ error at deeper levels (37.3\% vs.\ 96.2\% for $H^2SP$), where $H^2SP$ maintains 93--100\% diagonal recall. Concretely, on an unseen NTU60 test instance of \textit{writing} (A012), FLORA predicts the leaf, $L_3$, and $L_2$ labels correctly—which would recover all ancestors via tree lookup—yet matches its direct $L_1$ query to \textit{Interpersonal activity} instead of \textit{Information / signs}, illustrating the distinction between direct internal-description matching and lookup-based label recovery, whereas $H^2SP$ recovers the full path under both decoders.

\subsection{Ablations and Tree-Aware Euclidean Baselines}
\label{sec:abl}
We evaluate all variants on NTU60 55/5 and NTU120 96/24
(Fig.~\ref{fig:h2sp_ablation}).
Within each split, all variants use the same training and
pseudo-unseen validation classes, with checkpoints selected
by validation leaf Top-1 accuracy.
The loss-removal, no-composition and per-branch-gate variants retain all
other hyperparameters selected for full $H^2SP$, whereas
the Euclidean and Euclidean + Cone controls are tuned
independently using the same validation criterion.
Both controls retain the training taxonomy, shared-gate
composition, and semantic and distributional alignment losses.
Euclidean omits cone regularization, whereas Euclidean + Cone
uses a Euclidean cone penalty.
 
\par\addvspace{\intextsep}\noindent\begin{minipage}{\columnwidth}
    \centering
    \includegraphics[width=0.94\linewidth,trim=0 13bp 0 2.5bp,clip]{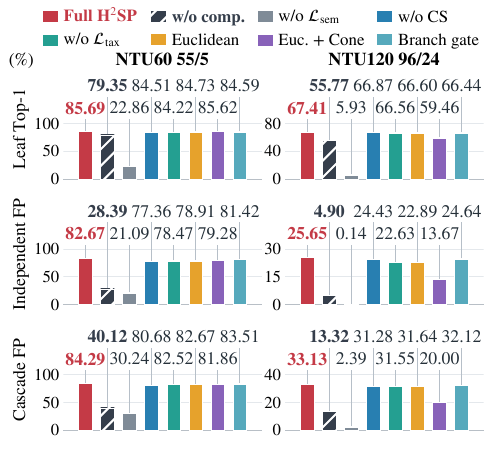}
\captionof{figure}{Component and geometry ablations. Columns correspond to NTU60 55/5 and NTU120 96/24; rows show leaf Top-1, independent and cascade full-path (FP) accuracy (\%). Legend: comp., parent–child composition; CS, Cauchy–Schwarz alignment; Euc., Euclidean; Branch gate, one gate per $L_1$ subtree; w/o, removal.}
\label{fig:h2sp_ablation}
\end{minipage}\par\addvspace{\intextsep}
 
Removing $\mathcal{L}_{\mathrm{sem}}$ sharply reduces all three metrics on both splits. On NTU120 96/24, leaf, independent full-path, and cascade full-path accuracies fall to 5.93\%, 0.14\%, and 2.39\%, respectively. Removing $\mathcal{L}_{\mathrm{CS}}$ lowers these metrics by 0.54, 1.22, and 1.85 percentage points; removing $\mathcal{L}_{\mathrm{tax}}$ lowers them by 0.85, 3.02, and 1.58 points. On both splits, the reductions in full-path accuracy exceed those in leaf accuracy for these two terms, indicating larger effects on hierarchical prediction. Their gains under independent decoding also indicate that the improvements are not solely due to cascade routing.
 
\noindent\textbf{Composition mechanism.}
Removing parent--child composition (Fig.~\ref{fig:h2sp_ablation}) sharply reduces leaf, independent, and cascade full-path accuracies from 85.69/82.67/84.29\% to 79.35/28.39/40.12\% on NTU60 55/5, and from 67.41/25.65/33.13\% to 55.77/4.90/13.32\% on NTU120 96/24, confirming its benefit for both leaf recognition and direct concept matching beyond cascade routing. Moreover, replacing the tree-wide gate with per-$L_1$-subtree scalars lowers these metrics by 1.10/1.25/0.78 and 0.97/1.01/1.01 points on the two splits, showing that a single shared gate improves accuracy without branch-specific parameters.

\noindent\textbf{Geometry and cone regularization.}
Without the cone, Lorentz and Euclidean full-path accuracies
differ by at most 0.44 points across the two splits.
Adding the cone in Lorentz space improves independent/cascade
full-path accuracy by 4.20/1.77 points on NTU60~55/5 and
3.02/1.58 points on NTU120~96/24.
Compared with Euclidean without cone regularization,
Euclidean + Cone differs by $+0.37/-0.81$ points on
NTU60~55/5 and $-9.23/-11.65$ points on NTU120~96/24
in independent/cascade full-path accuracy.
Full H$^{2}$SP exceeds Euclidean + Cone by 11.99/13.13
points on NTU120~96/24.
These results support the Lorentz-based formulation with
cone regularization in the evaluated settings.

\noindent\textbf{Prototype geometry.}
Fig.~\ref{fig:prototype_branches} shows all 127 composed prototypes of the NTU60
taxonomy $\mathcal T_A$, seen and unseen, from the full model of
Sec.~\ref{sec:abl}; each $L_1$ branch keeps its parent--child trajectories in
the same coordinates. Measured in the manifold itself (Fig.~\ref{fig:prototype_radius}),
the mean geodesic distance from the root grows with depth. Because composition and the cone constrain
parent--child order along each branch rather than imposing uniform shells,
distance ranges overlap across branches while every branch still expands
outward.

%% file: sec/tables/main_results_combined.tex
\newcommand{\bestresult}[1]{\textcolor{red}{\textbf{#1}}}
\newcommand{\secondresult}[1]{\textcolor{blue}{\textbf{#1}}}
\newcommand{\oursplain}[1]{\cellcolor{black!8}#1}
\newcommand{\oursbest}[1]{\cellcolor{black!8}\bestresult{#1}}
\newcommand{\ourssecond}[1]{\cellcolor{black!8}\secondresult{#1}}
\newcommand{\ourshead}{\cellcolor{black!8}H$^2$SP (Ours)}
\newcommand{\florahead}{FLORA (CVPR 2026)}
\newcommand{\fsvaehead}{FS-VAE (ICCV 2025)}
\newcommand{\tdsmhead}{TDSM (ICCV 2025)}
\newcommand{\sadvaehead}{SA-DVAE (ECCV 2024)}
\newcommand{\dvtahead}{DVTA (PR 2026)}
\newcommand{\smiehead}{SMIE (ACM MM 2023)}
\newcommand{\pgfahead}{PGFA$^\dagger$ (TIP 2025)}
\newcommand{\neuronhead}{Neuron$^\ddagger$ (CVPR 2025)}

\begin{table*}[!t]
\centering
\footnotesize
\setlength{\tabcolsep}{1.5pt}
\renewcommand{\arraystretch}{1}
\caption{Main comparison on the four NTU RGB+D splits. The first seven methods use the Standard Inductive Protocol, while the last two methods are shown under Modified / TTA Protocols. \bestresult{Red bold} and \secondresult{blue bold} values indicate the best and second-best results, respectively, among all displayed methods for each split and mean. $\dagger$ denotes a test-time adaptive configuration, and $\ddagger$ denotes a two-branch candidate-set hit rate on a filtered evaluation subset.}
\label{tab:main_results}
\begin{tabular*}{\textwidth}{@{\extracolsep{\fill}}ll*{15}{c}@{}}
\toprule
 & & \multicolumn{5}{c}{\textbf{\shortstack{(a) Independent full-path\\accuracy (\%)}}}
 & \multicolumn{5}{c}{\textbf{\shortstack{(b) Cascade full-path\\accuracy (\%)}}}
 & \multicolumn{5}{c}{\textbf{\shortstack{(c) Leaf-level Top-1\\accuracy (\%)}}}\\
\cmidrule(lr){3-7}\cmidrule(lr){8-12}\cmidrule(lr){13-17}
 & & \multicolumn{2}{c}{NTU60} & \multicolumn{2}{c}{NTU120} & \multirow{2}{*}{Mean}
 & \multicolumn{2}{c}{NTU60} & \multicolumn{2}{c}{NTU120} & \multirow{2}{*}{Mean}
 & \multicolumn{2}{c}{NTU60} & \multicolumn{2}{c}{NTU120} & \multirow{2}{*}{Mean}\\
\cmidrule(lr){3-4}\cmidrule(lr){5-6}\cmidrule(lr){8-9}\cmidrule(lr){10-11}\cmidrule(lr){13-14}\cmidrule(lr){15-16}
 & Method & 55/5 & 48/12 & 110/10 & 96/24 &
 & 55/5 & 48/12 & 110/10 & 96/24 &
 & 55/5 & 48/12 & 110/10 & 96/24 &\\
\midrule

\multirow{7}{*}{\textbf{\shortstack[l]{Standard\\Inductive\\Protocol}}} & \smiehead & 56.78 & 9.97 & 14.96 & 1.14 & 20.71 & 62.09 & 22.20 & 41.80 & 6.09 & 33.05 & 77.95 & 40.92 & 68.91 & 43.67 & 57.86 \\
 & \sadvaehead & 66.30 & 10.52 & 3.16 & 3.28 & 20.82 & 70.28 & 21.90 & 28.66 & 16.71 & 34.39 & 82.52 & 41.44 & 68.54 & 45.78 & 59.57 \\
 & \fsvaehead & 34.73 & 0.06 & 14.44 & 6.40 & 13.91 & 50.07 & 18.26 & 36.25 & 18.57 & 30.79 & 82.96 & 40.98 & 70.70 & 60.47 & 63.78 \\
 & \tdsmhead & 33.48 & 5.17 & 0.19 & 1.01 & 9.96 & 54.64 & 15.69 & 16.37 & 14.34 & 25.26 & 83.16 & 56.42 & 72.40 & \bestresult{67.84} & 69.96 \\
 & \dvtahead & 37.24 & 7.80 & 14.41 & 4.46 & 15.98 & 54.79 & 24.89 & \secondresult{44.16} & 14.22 & 34.52 & 77.80 & 37.55 & 71.55 & 47.57 & 58.62 \\
 & \florahead & 22.05 & 0.00 & 0.61 & 1.94 & 6.15 & 37.24 & 12.20 & 25.11 & 9.54 & 21.02 & \secondresult{86.28} & \bestresult{63.55} & \bestresult{79.30} & 64.03 & \bestresult{73.29} \\
 & \ourshead & \oursbest{82.67} & \oursbest{30.76} & \oursbest{34.76} & \oursbest{25.65} & \oursbest{43.46} & \oursbest{84.29} & \oursbest{39.76} & \oursbest{45.36} & \oursbest{33.13} & \oursbest{50.64} & \oursplain{85.69} & \oursplain{59.57} & \ourssecond{76.92} & \ourssecond{67.41} & \ourssecond{72.40} \\
\midrule

\multirow{2}{*}{\textbf{\shortstack[l]{Modified / TTA\\Protocols}}} & \neuronhead & 45.03 & \secondresult{15.07} & 17.37 & 2.72 & 20.05 & 64.47 & \secondresult{26.98} & 38.36 & 22.35 & 38.04 & \bestresult{87.50} & \secondresult{63.29} & 66.93 & 46.66 & 66.10 \\
 & \pgfahead & \secondresult{79.24} & 4.25 & \secondresult{25.30} & \secondresult{16.04} & \secondresult{31.21} & \secondresult{80.70} & 18.32 & 38.90 & \secondresult{23.80} & \secondresult{40.43} & 81.43 & 46.82 & 71.95 & 47.71 & 61.98 \\
\bottomrule
\end{tabular*}
\end{table*}

%% file: sec/tables/cross_taxonomy.tex
\begin{table}[!t]
\centering
\caption{\textbf{Cross-taxonomy transfer (\%)} from training
$\mathcal T_A$ to alternative $\mathcal T_B$.
Euc.\ + Cone denotes the Euclidean control with cone
regularization, independently tuned using the same
validation protocol as H$^2$SP.
Ind./Cas.: independent/cascade full-path accuracy.
\textbf{\textcolor{red}{Red}}/\textbf{\textcolor{blue}{blue}}
bold: best/second-best per split and metric.}
\label{tab:cross_tree}
\footnotesize
\setlength{\tabcolsep}{1.7pt}
\begin{tabular*}{\columnwidth}{@{\extracolsep{\fill}}l*{8}{r}@{}}
\toprule
& \multicolumn{2}{c}{\textbf{H$^2$SP (Ours)}}
& \multicolumn{2}{c}{Euc.\ + Cone}
& \multicolumn{2}{c}{FLORA}
& \multicolumn{2}{c}{PGFA$^\dagger$} \\
\cmidrule(lr){2-3}\cmidrule(lr){4-5}\cmidrule(lr){6-7}\cmidrule(lr){8-9}
Split & Ind. & Cas. & Ind. & Cas. & Ind. & Cas. & Ind. & Cas. \\
\midrule
NTU60 55/5
& \textbf{\textcolor{red}{21.39}} & \textbf{\textcolor{red}{29.87}}
& \textbf{\textcolor{blue}{19.47}} & 21.42
& 2.88 & 19.32
& 0.22 & \textbf{\textcolor{blue}{21.56}} \\
NTU60 48/12
& \textbf{\textcolor{red}{9.08}} & \textbf{\textcolor{red}{13.33}}
& \textbf{\textcolor{blue}{7.81}} & \textbf{\textcolor{blue}{11.79}}
& 3.18 & 10.15
& 0.09 & 11.39 \\
NTU120 110/10
& \textbf{\textcolor{red}{11.61}} & \textbf{\textcolor{red}{15.00}}
& \textbf{\textcolor{blue}{10.66}} & \textbf{\textcolor{blue}{14.96}}
& 1.74 & 10.74
& 4.21 & 6.57 \\
NTU120 96/24
& \textbf{\textcolor{red}{4.60}} & \textbf{\textcolor{red}{11.41}}
& \textbf{\textcolor{blue}{1.72}} & 4.14
& 1.55 & \textbf{\textcolor{blue}{4.34}}
& 0.27 & 3.86 \\
\midrule
Mean
& \textbf{\textcolor{red}{11.67}} & \textbf{\textcolor{red}{17.40}}
& \textbf{\textcolor{blue}{9.92}} & \textbf{\textcolor{blue}{13.08}}
& 2.34 & 11.14
& 1.20 & 10.85 \\
\bottomrule
\end{tabular*}
\end{table}

%% file: sec/6_conclusion.tex
\section{Conclusion}
\label{sec:concl}
We studied whether zero-shot skeleton representations recognize broader concepts beyond leaf labels, showing that strong leaf classification alone does not ensure direct ancestor matching. By composing Lorentz prototypes via shared-gate geodesic interpolation with seen-leaf alignment and soft entailment cones, $H^2SP$ achieves the best mean independent and cascade full-path accuracies across four NTU RGB+D splits with competitive leaf accuracy. Its gains persist under zero-retraining transfer to an alternative taxonomy, and ablations support shared gating and Lorentz-cone coupling. Improving recognition of unseen internal concepts remains future work. Our results identify direct internal-concept matching as a complementary diagnostic of motion grounding and hyperbolic hierarchical prototypes as an effective approach across granularities.

%% file: main.bib
@String(CVPR= {IEEE Conf. Comput. Vis. Pattern Recog.})

@String(ICCV= {Int. Conf. Comput. Vis.})

@String(AAAI = {AAAI})

@String(CVPR  = {CVPR})

@String(ICCV  = {ICCV})

@inproceedings{flora,
  title={Learning by neighbor-aware semantics, deciding by open-form flows: Towards robust zero-shot skeleton action recognition},
  author={Chen, Yang and Li, Miaoge and Rao, Zhijie and Zeng, Deze and Guo, Song and Guo, Jingcai},
  booktitle={Proceedings of the IEEE/CVF Conference on Computer Vision and Pattern Recognition},
  pages={3374--3383},
  year={2026}
}

@inproceedings{tdsm,
  title={Bridging the skeleton-text modality gap: diffusion-powered modality alignment for zero-shot skeleton-based action recognition},
  author={Do, Jeonghyeok and Kim, Munchurl},
  booktitle={2025 IEEE/CVF International Conference on Computer Vision (ICCV)},
  pages={12757--12768},
  year={2025},
  organization={IEEE}
}

@inproceedings{fsvae,
  title={Frequency-semantic enhanced variational autoencoder for zero-shot skeleton-based action recognition},
  author={Wu, Wenhan and Guo, Zhishuai and Chen, Chen and Xue, Hongfei and Lu, Aidong},
  booktitle={2025 IEEE/CVF International Conference on Computer Vision (ICCV)},
  pages={11122--11131},
  year={2025},
  organization={IEEE}
}

@inproceedings{neuron,
  title={Neuron: Learning context-aware evolving representations for zero-shot skeleton action recognition},
  author={Chen, Yang and Guo, Jingcai and Guo, Song and Tao, Dacheng},
  booktitle={2025 IEEE/CVF Conference on Computer Vision and Pattern Recognition (CVPR)},
  pages={8721--8730},
  year={2025},
  organization={IEEE}
}

@inproceedings{clip,
  title={Learning transferable visual models from natural language supervision},
  author={Radford, Alec and Kim, Jong Wook and Hallacy, Chris and Ramesh, Aditya and Goh, Gabriel and Agarwal, Sandhini and Sastry, Girish and Askell, Amanda and Mishkin, Pamela and Clark, Jack and others},
  booktitle={International conference on machine learning},
  pages={8748--8763},
  year={2021},
  organization={PMLR}
}

@inproceedings{lorentz,
  title={Learning continuous hierarchies in the lorentz model of hyperbolic geometry},
  author={Nickel, Maximillian and Kiela, Douwe},
  booktitle={International conference on machine learning},
  pages={3779--3788},
  year={2018},
  organization={PMLR}
}

@inproceedings{cones,
  title={Hyperbolic entailment cones for learning hierarchical embeddings},
  author={Ganea, Octavian and B{\'e}cigneul, Gary and Hofmann, Thomas},
  booktitle={International conference on machine learning},
  pages={1646--1655},
  year={2018},
  organization={PMLR}
}

@inproceedings{meru,
  title={Hyperbolic image-text representations},
  author={Desai, Karan and Nickel, Maximilian and Rajpurohit, Tanmay and Johnson, Justin and Vedantam, Shanmukha Ramakrishna},
  booktitle={International Conference on Machine Learning},
  pages={7694--7731},
  year={2023},
  organization={PMLR}
}

@inproceedings{ntu60,
  title={Ntu rgb+ d: A large scale dataset for 3d human activity analysis},
  author={Shahroudy, Amir and Liu, Jun and Ng, Tian-Tsong and Wang, Gang},
  booktitle={Proceedings of the IEEE conference on computer vision and pattern recognition},
  pages={1010--1019},
  year={2016}
}

@inproceedings{shiftgcn,
  title={Skeleton-based action recognition with shift graph convolutional network},
  author={Cheng, Ke and Zhang, Yifan and He, Xiangyu and Chen, Weihan and Cheng, Jian and Lu, Hanqing},
  booktitle={2020 IEEE/CVF Conference on Computer Vision and Pattern Recognition (CVPR)},
  pages={180--189},
  year={2020},
  organization={IEEE}
}

@ARTICLE{ntu120,
  author={Liu, Jun and Shahroudy, Amir and Perez, Mauricio and Wang, Gang and Duan, Ling-Yu and Kot, Alex C.},
  journal={IEEE Transactions on Pattern Analysis and Machine Intelligence}, 
  title={NTU RGB+D 120: A Large-Scale Benchmark for 3D Human Activity Understanding}, 
  year={2020},
  volume={42},
  number={10},
  pages={2684-2701},
  doi={10.1109/TPAMI.2019.2916873}}

@article{pgfa,
  title={Zero-shot skeleton-based action recognition with prototype-guided feature alignment},
  author={Zhou, Kai and Zhang, Shuhai and You, Zeng and Hu, Jinwu and Tan, Mingkui and Liu, Fei},
  journal={IEEE Transactions on Image Processing},
  volume={34},
  pages={4602--4617},
  year={2025},
  publisher={IEEE}
}

@inproceedings{protofm,
  title={Beyond Alignment: A Generative Matching Paradigm via Flow Matching for Zero-Shot Skeleton-Based Action Recognition},
  author={Liu, Xuan and Wu, Cong and Fang, Wei and Feng, Zhenhua},
  booktitle={European Conference on Computer Vision},
  pages={1--18},
  year={2026},
  organization={Springer}
}

@inproceedings{sadvae,
  title={Sa-dvae: Improving zero-shot skeleton-based action recognition by disentangled variational autoencoders},
  author={Li, Sheng-Wei and Wei, Zi-Xiang and Chen, Wei-Jie and Yu, Yi-Hsin and Yang, Chih-Yuan and Hsu, Jane Yung-jen},
  booktitle={European conference on computer vision},
  pages={447--462},
  year={2024},
  organization={Springer}
}

@inproceedings{smie,
  title={Zero-shot skeleton-based action recognition via mutual information estimation and maximization},
  author={Zhou, Yujie and Qiang, Wenwen and Rao, Anyi and Lin, Ning and Su, Bing and Wang, Jiaqi},
  booktitle={Proceedings of the 31st ACM international conference on multimedia},
  pages={5302--5310},
  year={2023}
}

@article{dvta,
    title   = {Zero-shot skeleton-based action recognition with dual visual-text alignment},
      author  = {Jidong Kuang and Hongsong Wang and Chaolei Han and Yang Zhang and Jie Gui},
      journal = {Pattern Recognition},
      volume  = {171},
      pages   = {112342},
      year    = {2026},
      issn    = {0031-3203}
}

@article{li2025learning,
  title={{H$^2$em}: Learning Hierarchical Hyperbolic Embeddings for Compositional Zero-Shot Learning},
  author={Li, Lin and Li, Jiahui and Lei, Jiaming and Xiao, Jun and Shao, Feifei and Chen, Long},
  journal={arXiv preprint arXiv:2512.20029},
  year={2025}
}

@inproceedings{csaligner,
  title={Distributional vision-language alignment by cauchy-schwarz divergence},
  author={Yin, Wenzhe and Xiao, Zehao and Zhou, Pan and Yu, Shujian and Shen, Jiayi and Sonke, Jan-Jakob and Gavves, Efstratios},
  booktitle={International Conference on Learning Representations},
  volume={2026},
  pages={39688--39710},
  year={2026}
}

@article{gap,
  title={Is the Modality Gap a Bug or a Feature? A Robustness Perspective},
  author={Chowers, Rhea and Naparstek, Oshri and Barzelay, Udi and Weiss, Yair},
  journal={arXiv preprint arXiv:2603.29080},
  year={2026}
}

@article{dynapurls,
  title={DynaPURLS: Dynamic Refinement of Part-Aware Representations for Skeleton-Based Zero-Shot Action Recognition},
  author={Zhu, Jingmin and Zhu, Anqi and Bailey, James and Liu, Jun and Rahmani, Hossein and Bennamoun, Mohammed and Boussaid, Farid and Ke, Qiuhong},
  journal={IEEE Transactions on Pattern Analysis and Machine Intelligence},
  year={2026},
  publisher={IEEE}
}

@inproceedings{skelhcc,
  title={{SkelHCC}: A Hyperbolic {CLIP}-Driven Cache Adaptation Framework for Skeleton-based One-Shot Action Recognition},
  author={Liu, Yanan and Zhu, Anqi and Zhu, Jingmin and Liu, Jun and Rahmani, Hossein and Bennamoun, Mohammed and Boussaid, Farid and Xu, Dan and Ke, Qiuhong},
  booktitle={International Conference on Machine Learning},
  year={2026}
}

@inproceedings{pangea,
  title={From isolated islands to pangea: Unifying semantic space for human action understanding},
  author={Li, Yong-Lu and Wu, Xiaoqian and Liu, Xinpeng and Wang, Zehao and Dou, Yiming and Ji, Yikun and Zhang, Junyi and Li, Yixing and Lu, Xudong and Tan, Jingru and others},
  booktitle={2024 IEEE/CVF Conference on Computer Vision and Pattern Recognition (CVPR)},
  pages={16582--16592},
  year={2024},
  organization={IEEE}
}

@article{universal,
  title={Toward Universal Skeleton-Based Action Recognition across Heterogeneous Skeletons and Open Vocabularies},
  author={Kuang, Jidong and Wang, Hongsong and Gui, Jie and Tang, Yuan Yan and Kwok, James Tin-Yau},
  journal={arXiv preprint arXiv:2604.17013},
  year={2026}
}

@article{attention,
  title={Attention is all you need},
  author={Vaswani, Ashish and Shazeer, Noam and Parmar, Niki and Uszkoreit, Jakob and Jones, Llion and Gomez, Aidan N and Kaiser, {\L}ukasz and Polosukhin, Illia},
  journal={Advances in neural information processing systems},
  volume={30},
  year={2017}
}

@article{cpc,
  title={Representation learning with contrastive predictive coding},
  author={Oord, Aaron van den and Li, Yazhe and Vinyals, Oriol},
  journal={arXiv preprint arXiv:1807.03748},
  year={2018}
}

@inproceedings{hyperbolic_action,
  title={Searching for Actions on the Hyperbole},
  author={Long, Teng and Mettes, Pascal and Shen, Heng Tao and Snoek, Cees G. M.},
  booktitle={2020 IEEE/CVF Conference on Computer Vision and Pattern Recognition (CVPR)},
  pages={1141--1150},
  year={2020},
  organization={IEEE}
}

@inproceedings{stgcn,
  title     = {Spatial Temporal Graph Convolutional Networks for Skeleton-Based Action Recognition},
  author    = {Yan, Sijie and Xiong, Yuanjun and Lin, Dahua},
  booktitle = {Proceedings of the AAAI Conference on Artificial Intelligence},
  volume    = {32},
  year      = {2018}
}

@inproceedings{2sagcn,
  title     = {Two-Stream Adaptive Graph Convolutional Networks for Skeleton-Based Action Recognition},
  author    = {Shi, Lei and Zhang, Yifan and Cheng, Jian and Lu, Hanqing},
  booktitle = {Proceedings of the IEEE/CVF Conference on Computer Vision and Pattern Recognition},
  pages     = {12026--12035},
  year      = {2019}
}

@inproceedings{msg3d,
  title     = {Disentangling and Unifying Graph Convolutions for Skeleton-Based Action Recognition},
  author    = {Liu, Ziyu and Zhang, Hongwen and Chen, Zhenghao and Wang, Zhiyong and Ouyang, Wanli},
  booktitle = {Proceedings of the IEEE/CVF Conference on Computer Vision and Pattern Recognition},
  pages     = {143--152},
  year      = {2020}
}

@inproceedings{ctrgcn,
  title     = {Channel-Wise Topology Refinement Graph Convolution for Skeleton-Based Action Recognition},
  author    = {Chen, Yuxin and Zhang, Ziqi and Yuan, Chunfeng and Li, Bing and Deng, Ying and Hu, Weiming},
  booktitle = {Proceedings of the IEEE/CVF International Conference on Computer Vision},
  pages     = {13359--13368},
  year      = {2021}
}

@inproceedings{infogcn,
  title     = {{InfoGCN}: Representation Learning for Human Skeleton-Based Action Recognition},
  author    = {Chi, Hyung-gun and Ha, Myoung Hoon and Chi, Seunggeun and Lee, Sang Wan and Huang, Qixing and Ramani, Karthik},
  booktitle = {Proceedings of the IEEE/CVF Conference on Computer Vision and Pattern Recognition},
  pages     = {20186--20196},
  year      = {2022}
}

@inproceedings{posec3d,
  title     = {Revisiting Skeleton-Based Action Recognition},
  author    = {Duan, Haodong and Zhao, Yue and Chen, Kai and Lin, Dahua and Dai, Bo},
  booktitle = {Proceedings of the IEEE/CVF Conference on Computer Vision and Pattern Recognition},
  pages     = {2969--2978},
  year      = {2022}
}

@inproceedings{hdgcn,
  title     = {Hierarchically Decomposed Graph Convolutional Networks for Skeleton-Based Action Recognition},
  author    = {Lee, Jungho and Lee, Minhyeok and Lee, Dogyoon and Lee, Sangyoun},
  booktitle = {Proceedings of the IEEE/CVF International Conference on Computer Vision},
  pages     = {10444--10453},
  year      = {2023}
}

@inproceedings{lst,
  title     = {Generative Action Description Prompts for Skeleton-based Action Recognition},
  author    = {Xiang, Wangmeng and Li, Chao and Zhou, Yuxuan and Wang, Biao and Zhang, Lei},
  booktitle = {Proceedings of the IEEE/CVF International Conference on Computer Vision},
  pages     = {10276--10285},
  year      = {2023}
}

@inproceedings{synse,
  title     = {Syntactically Guided Generative Embeddings for Zero-Shot Skeleton Action Recognition},
  author    = {Gupta, Pranay and Sharma, Divyanshu and Sarvadevabhatla, Ravi Kiran},
  booktitle = {Proceedings of the IEEE International Conference on Image Processing},
  pages     = {439--443},
  year      = {2021}
}

@article{jpose,
  title   = {Skeleton based Zero Shot Action Recognition in Joint Pose-Language Semantic Space},
  author  = {Jasani, Bhavan and Mazagonwalla, Afshaan},
  journal = {arXiv preprint arXiv:1911.11344},
  year    = {2019}
}

@inproceedings{purls,
  title     = {Part-aware Unified Representation of Language and Skeleton for Zero-shot Action Recognition},
  author    = {Zhu, Anqi and Ke, Qiuhong and Gong, Mingming and Bailey, James},
  booktitle = {Proceedings of the IEEE/CVF Conference on Computer Vision and Pattern Recognition},
  pages     = {18761--18770},
  year      = {2024}
}

@inproceedings{star,
  title     = {Fine-Grained Side Information Guided Dual-Prompts for Zero-Shot Skeleton Action Recognition},
  author    = {Chen, Yang and Guo, Jingcai and He, Tian and Lu, Xiaocheng and Wang, Ling},
  booktitle = {Proceedings of the 32nd ACM International Conference on Multimedia},
  pages     = {778--786},
  year      = {2024}
}

@article{infocomp,
  title   = {An Information Compensation Framework for Zero-Shot Skeleton-Based Action Recognition},
  author  = {Xu, Haojun and Gao, Yan and Li, Jie and Gao, Xinbo},
  journal = {IEEE Transactions on Multimedia},
  volume  = {27},
  pages   = {4882--4894},
  year    = {2025}
}

@inproceedings{msf,
  title     = {Multi-semantic Fusion Model for Generalized Zero-Shot Skeleton-Based Action Recognition},
  author    = {Li, Ming-Zhe and Jia, Zhen and Zhang, Zhang and Ma, Zhanyu and Wang, Liang},
  booktitle = {International Conference on Image and Graphics},
  pages     = {68--80},
  year      = {2023},
  publisher = {Springer}
}

@inproceedings{devise,
  title     = {{DeViSE}: A Deep Visual-Semantic Embedding Model},
  author    = {Frome, Andrea and Corrado, Greg S. and Shlens, Jon and Bengio, Samy and Dean, Jeff and Ranzato, Marc'Aurelio and Mikolov, Tomas},
  booktitle = {Advances in Neural Information Processing Systems},
  volume    = {26},
  year      = {2013}
}

@article{labelembed,
  title   = {Label-Embedding for Image Classification},
  author  = {Akata, Zeynep and Perronnin, Florent and Harchaoui, Zaid and Schmid, Cordelia},
  journal = {IEEE Transactions on Pattern Analysis and Machine Intelligence},
  volume  = {38},
  number  = {7},
  pages   = {1425--1438},
  year    = {2016}
}

@inproceedings{rohrbach,
  title     = {What Helps Where -- And Why? Semantic Relatedness for Knowledge Transfer},
  author    = {Rohrbach, Marcus and Stark, Michael and Szarvas, Gy{\"o}rgy and Gurevych, Iryna and Schiele, Bernt},
  booktitle = {Proceedings of the IEEE Computer Society Conference on Computer Vision and Pattern Recognition},
  pages     = {910--917},
  year      = {2010}
}

@inproceedings{hierloss,
  title     = {Making Better Mistakes: Leveraging Class Hierarchies With Deep Networks},
  author    = {Bertinetto, Luca and Mueller, Romain and Tertikas, Konstantinos and Samangooei, Sina and Lord, Nicholas A.},
  booktitle = {Proceedings of the IEEE/CVF Conference on Computer Vision and Pattern Recognition},
  pages     = {12506--12515},
  year      = {2020}
}

@inproceedings{finegym,
  title     = {{FineGym}: A Hierarchical Video Dataset for Fine-Grained Action Understanding},
  author    = {Shao, Dian and Zhao, Yue and Dai, Bo and Lin, Dahua},
  booktitle = {Proceedings of the IEEE/CVF Conference on Computer Vision and Pattern Recognition},
  pages     = {2616--2625},
  year      = {2020}
}

@inproceedings{hiod,
  title     = {Hierarchical-aware Orthogonal Disentanglement Framework for Fine-grained Skeleton-based Action Recognition},
  author    = {Chang, Haochen and Ren, Pengfei and Zhang, Haoyang and Xie, Liang and Chen, Hongbo and Yin, Erwei},
  booktitle = {Proceedings of the IEEE/CVF International Conference on Computer Vision},
  pages     = {11252--11261},
  year      = {2025}
}

@inproceedings{poincare,
  title     = {{P}oincar{\'e} Embeddings for Learning Hierarchical Representations},
  author    = {Nickel, Maximilian and Kiela, Douwe},
  booktitle = {Advances in Neural Information Processing Systems},
  volume    = {30},
  year      = {2017}
}

@inproceedings{lorentzdist,
  title     = {{L}orentzian Distance Learning for Hyperbolic Representations},
  author    = {Law, Marc and Liao, Renjie and Snell, Jake and Zemel, Richard},
  booktitle = {International Conference on Machine Learning},
  pages     = {3672--3681},
  year      = {2019}
}

@inproceedings{vendrov,
  title     = {Order-Embeddings of Images and Language},
  author    = {Vendrov, Ivan and Kiros, Ryan and Fidler, Sanja and Urtasun, Raquel},
  booktitle = {International Conference on Learning Representations},
  year      = {2016}
}

@inproceedings{hnn,
  title     = {Hyperbolic Neural Networks},
  author    = {Ganea, Octavian and B{\'e}cigneul, Gary and Hofmann, Thomas},
  booktitle = {Advances in Neural Information Processing Systems},
  volume    = {31},
  year      = {2018}
}

@inproceedings{hgcn,
  title     = {Hyperbolic Graph Convolutional Neural Networks},
  author    = {Chami, Ines and Ying, Zhitao and R{\'e}, Christopher and Leskovec, Jure},
  booktitle = {Advances in Neural Information Processing Systems},
  volume    = {32},
  year      = {2019}
}

@inproceedings{fullyhyp,
  title     = {Fully Hyperbolic Neural Networks},
  author    = {Chen, Weize and Han, Xu and Lin, Yankai and Zhao, Hexu and Liu, Zhiyuan and Li, Peng and Sun, Maosong and Zhou, Jie},
  booktitle = {Proceedings of the 60th Annual Meeting of the Association for Computational Linguistics (Volume 1: Long Papers)},
  pages     = {5672--5686},
  year      = {2022}
}

@inproceedings{khrulkov,
  title     = {Hyperbolic Image Embeddings},
  author    = {Khrulkov, Valentin and Mirvakhabova, Leyla and Ustinova, Evgeniya and Oseledets, Ivan and Lempitsky, Victor},
  booktitle = {Proceedings of the IEEE/CVF Conference on Computer Vision and Pattern Recognition},
  pages     = {6418--6428},
  year      = {2020}
}

@inproceedings{hyptransfer,
  title     = {Hyperbolic Vision Transformers: Combining Improvements in Metric Learning},
  author    = {Ermolov, Aleksandr and Mirvakhabova, Leyla and Khrulkov, Valentin and Sebe, Nicu and Oseledets, Ivan},
  booktitle = {Proceedings of the IEEE/CVF Conference on Computer Vision and Pattern Recognition},
  pages     = {7409--7419},
  year      = {2022}
}

@inproceedings{atigh,
  title     = {Hyperbolic Image Segmentation},
  author    = {{Ghadimi Atigh}, Mina and Schoep, Julian and Acar, Erman and {van Noord}, Nanne and Mettes, Pascal},
  booktitle = {Proceedings of the IEEE/CVF Conference on Computer Vision and Pattern Recognition},
  pages     = {4453--4462},
  year      = {2022}
}

@inproceedings{hycoclip,
  title     = {Compositional Entailment Learning for Hyperbolic Vision-Language Models},
  author    = {Pal, Avik and {van Spengler}, Max and {D'Amely di Melendugno}, Guido Maria and Flaborea, Alessandro and Galasso, Fabio and Mettes, Pascal},
  booktitle = {International Conference on Learning Representations},
  year      = {2025}
}

@inproceedings{hyperfuture,
  title     = {Learning the Predictability of the Future},
  author    = {Sur{\'\i}s, D{\'\i}dac and Liu, Ruoshi and Vondrick, Carl},
  booktitle = {Proceedings of the IEEE/CVF Conference on Computer Vision and Pattern Recognition},
  pages     = {12607--12617},
  year      = {2021}
}

@inproceedings{hysp,
  title     = {Hyperbolic Self-paced Learning for Self-supervised Skeleton-based Action Representations},
  author    = {Franco, Luca and Mandica, Paolo and Munjal, Bharti and Galasso, Fabio},
  booktitle = {International Conference on Learning Representations},
  year      = {2023}
}

@article{zacks2001event,
  title={Event structure in perception and conception},
  author={Zacks, Jeffrey M. and Tversky, Barbara},
  journal={Psychological Bulletin},
  volume={127},
  number={1},
  pages={3--21},
  year={2001},
  publisher={American Psychological Association},
  doi={10.1037/0033-2909.127.1.3}
}

@article{aksoy2011learning,
  title={Learning the semantics of object--action relations by observation},
  author={Aksoy, Eren Erdal and Abramov, Alexey and D{\"o}rr, Johannes and Ning, Kejun and Dellen, Babette and W{\"o}rg{\"o}tter, Florentin},
  journal={The International Journal of Robotics Research},
  volume={30},
  number={10},
  pages={1229--1249},
  year={2011},
  publisher={Sage Publications, Sage UK: London, England}
}

@article{kipper2008verbnet,
  title={A large-scale classification of English verbs},
  author={Kipper, Karin and Korhonen, Anna and Ryant, Neville and Palmer, Martha},
  journal={Language Resources and Evaluation},
  volume={42},
  number={1},
  pages={21--40},
  year={2008},
  publisher={Springer}
}

@inproceedings{baker1998berkeley,
  title={The berkeley framenet project},
  author={Baker, Collin F and Fillmore, Charles J and Lowe, John B},
  booktitle={COLING 1998 Volume 1: The 17th International Conference on Computational Linguistics},
  year={1998}
}

@book{bach1968universals,
  title={Universals in Linguistic Theory},
  editor={Bach, Emmon and Harms, Robert T.},
  publisher={Holt, Rinehart and Winston},
  address={New York},
  year={1968}
}
